\documentclass{article}

\usepackage[preprint]{neurips_2026}
\usepackage{graphicx}
\usepackage{xcolor}
\usepackage{float}
\usepackage{subcaption}
\usepackage{booktabs}

\usepackage{placeins} 

\usepackage[utf8]{inputenc} 
\usepackage[T1]{fontenc}    
\usepackage{hyperref}       
\usepackage{url}            
\usepackage{amsfonts}       
\usepackage{nicefrac}       
\usepackage{microtype}      
\usepackage{amsmath}
\hypersetup{
    colorlinks=true,
    citecolor=blue,
    linkcolor=blue,
    urlcolor=blue
} 

\title{Structure over Depth: A Single-Block Spatio-Temporal Transformer for Multi-Entity Reasoning}

\author{%
\begin{tabular}{ccc}
Narthana Sivalingam &
Santhirarajah Sivasthigan &
Buddhi Wijenayake
\\[0.8ex]
Roshan Godaliyadda &
Vijitha Herath &
Parakrama Ekanayake
\end{tabular}
\\[1.5ex]
Department of Electrical and Electronic Engineering \\
University of Peradeniya \\
Peradeniya, Sri Lanka \\
}

\begin{document}

\maketitle

\begin{abstract}
Modeling multi-entity temporal data requires capturing complex dependencies across entities, across time, and across their interactions. While transformer-based approaches have shown strong performance, they typically rely on deep stacking of transformer layers to implicitly learn these heterogeneous dependencies, leading to high computational cost. In this work, we revisit this problem from a structural perspective and show that multi-entity temporal dynamics can be decomposed into three fundamental interaction types: spatial interactions among entities, temporal interactions across time, and cross interactions that couple these two domains. Motivated by this observation, we propose a structured spatio-temporal transformer block that explicitly and jointly models these interaction types within a single stage. The architecture consists of parallel spatial and temporal self-attention modules, followed by a bidirectional cross-attention mechanism, with outputs combined through a learnable gated fusion. By directly encoding the three core interaction views, the proposed design substantially reduces reliance on deep stacking to approximate these dependencies. We validate our approach across diverse domains, including video-based group activity recognition, skeleton-based human interaction analysis, and wearable sensor-based activity recognition. Despite its simplicity, the proposed single structured Transformer block consistently matches or outperforms deeper transformer architectures with just only 1.76M parameters. Our findings suggest that the necessity of depth in prior models arises from implicit and entangled modeling, and demonstrate that explicit interaction factorization provides a more efficient alternative while improving transparency in how different interaction types are captured. More broadly, this work points to structure-first design: expressive, efficient, and transparent multi-entity temporal reasoning can emerge by exposing interaction structure rather than relying on depth.
\end{abstract}

\section{INTRODUCTION}

\begin{figure}[t]
    \centering

    \begin{subfigure}{0.3\linewidth}
        \centering
        \includegraphics[width=\linewidth]{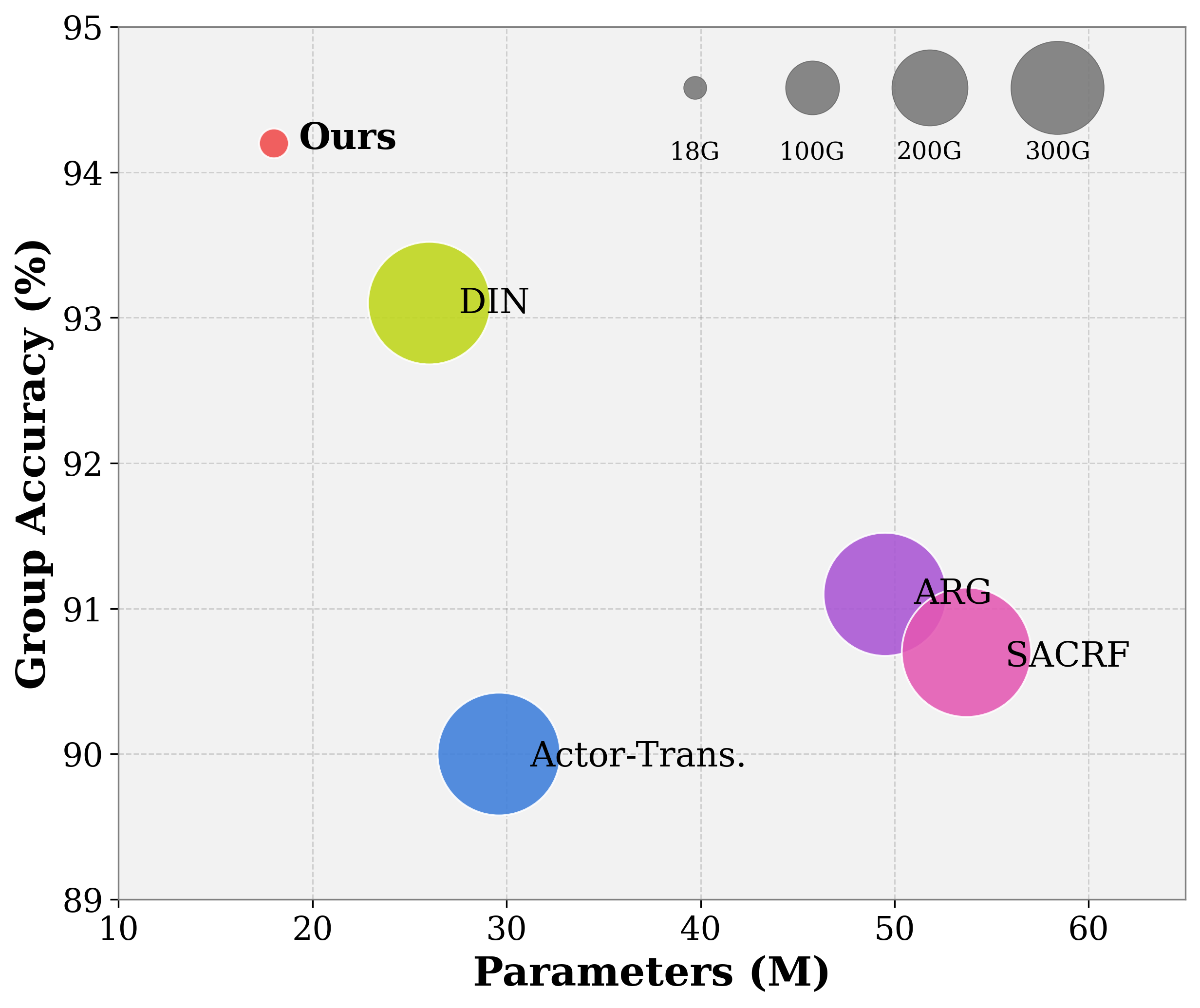}
        \caption{Volleyball dataset.}
        \label{fig:intro_bubble_volleyball}
    \end{subfigure}
    \hfill
    \begin{subfigure}{0.3\linewidth}
        \centering
        \includegraphics[width=\linewidth]{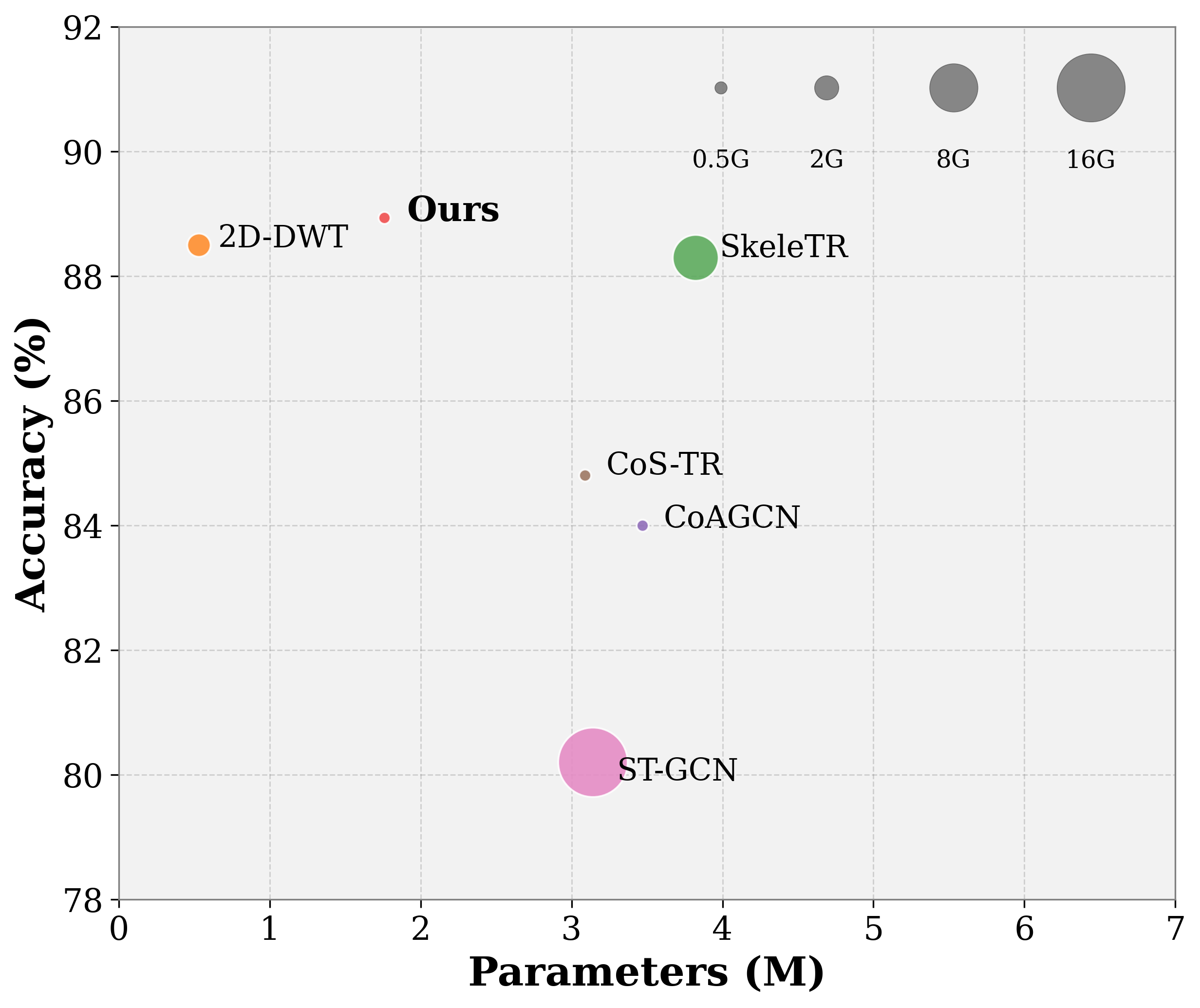}
        \caption{NTU120 cross-subject.}
        \label{fig:intro_bubble_ntu120}
    \end{subfigure}
\hfill
    \begin{subfigure}{0.3\linewidth}
        \centering
        \includegraphics[width=\linewidth]{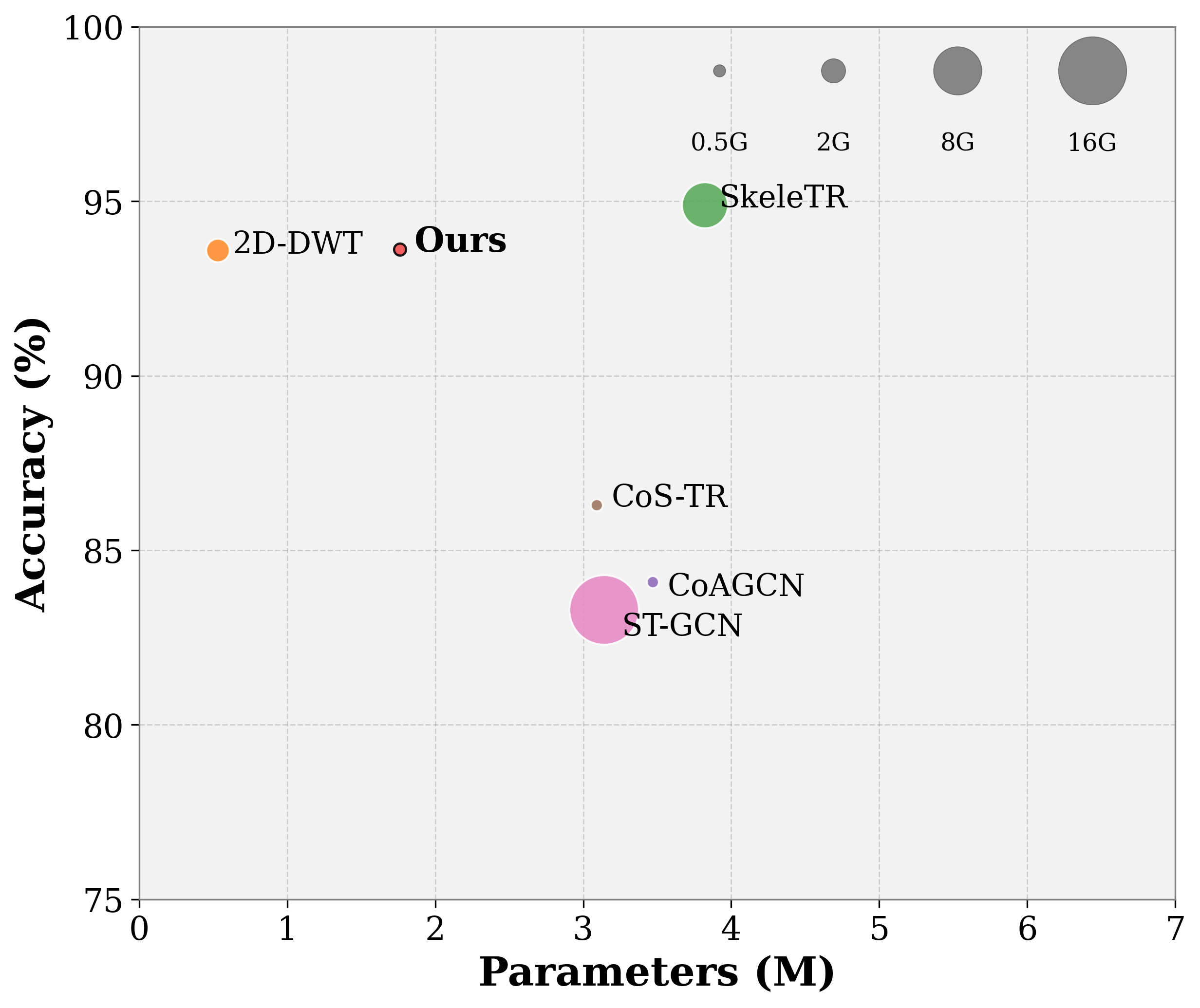}
        \caption{NTU60 cross-subject.}
        \label{fig:intro_bubble_ntu120}
    \end{subfigure}
    \caption{
    Representative accuracy-efficiency trade-off. The x-axis denotes parameters, the y-axis denotes accuracy, and bubble size denotes FLOPs. Our proposed model achieves a competitive high-accuracy, low-parameter, and low-computation position.
    }
    \label{fig:intro_accuracy_efficiency}
    \vspace{-0.8em}
\end{figure}

Understanding collective behavior from multi-entity temporal data is a fundamental challenge in modern machine learning. Tasks such as video-based group activity recognition, skeleton-based human--human interaction understanding, and wearable sensor-based activity recognition require reasoning over multiple entities evolving over time \cite{ibrahim2016hierarchical,NTU_DATASET,liu2020ntu120}. Unlike single-entity recognition, these settings demand modeling dependencies across entities, across time, and across their interactions. Such reasoning is critical for surveillance, sports analytics, healthcare monitoring, autonomous systems, human--robot interaction, and wearable intelligence. Yet many of these applications operate under strict constraints: limited memory, limited computation, low power budgets, and fast inference on edge or embedded devices. Thus, multi-entity temporal reasoning is not only a question of accuracy, but also of efficiency, deployability, and structural simplicity.

Existing approaches have evolved from CNN/RNN architectures to graph-based and Transformer-based relational models. CNNs and RNNs learn appearance and temporal patterns effectively, but provide limited mechanisms for explicit multi-entity coordination\cite{carreira2017quo,simonyan2014twostream}. Graph-based methods encode structure through entity nodes and relational edges, yet often depend on predefined connectivity or modality-specific assumptions\cite{yan2018stgcn,wu2019arg}. Transformers avoid fixed graphs using data-dependent attention, but many spatio-temporal models gain relational capacity through deep stacks of generic Transformer layers\cite{vaswani2017attention, dosovitskiy2021vit}. This creates a deployment-relevant tension: depth improves expressiveness, but increases parameters, computation, memory, and latency. These costs are restrictive for edge-oriented and real-time applications, where models must capture spatial relations, temporal evolution, and coordinated multi-entity behavior while remaining accurate and lightweight.

This raises a central question: how much depth is truly required for multi-entity spatio-temporal reasoning, and how much can be replaced by explicit interaction structure? We argue that deep spatio-temporal Transformers often incur high cost because generic transformer layer stacks must learn heterogeneous dependencies jointly: entity interactions, temporal evolution, and spatio-temporal co-ordination between entities. While depth increases expressiveness, repeatedly mixing these dependencies inside generic computations can require larger capacity and produce entangled representations\cite{gavrilyuk2020actortransformer, li2021groupformer}.

We therefore revisit the problem from a structural perspective and decompose multi-entity spatio-temporal dynamics into three fundamental interaction types: spatial interactions among entities within a time step, temporal interactions across time for each entity, and cross spatio-temporal interactions that connect entity relationships with temporal dynamics. This suggests a structure-first design principle: rather than increasing depth to discover these interactions implicitly, a model can directly represent and combine the essential interaction views within a single structured stage.

Based on this principle, we propose a unified structured spatio-temporal Transformer framework. Raw observations, such as person-level visual features, skeleton sequences, or wearable sensor streams, are first converted into entity-level temporal representations. A single structured relational block then explicitly models the three interaction views: spatial self-attention captures relations among entities within each time step, temporal self-attention captures each entity's evolution over time, and bidirectional spatio-temporal cross-attention couples the two representations. Learnable gated fusion adaptively combines spatial, temporal, and cross-interaction features. This design bridges graph-based structural bias and Transformer flexibility: it avoids fixed hand-designed graphs while reducing reliance on deep generic attention stacks to learn all relations implicitly. Fig.~\ref{fig:intro_accuracy_efficiency} summarizes the resulting accuracy--efficiency trade-off on representative video and skeleton benchmarks. The proposed model lies in the favorable high-accuracy, low-parameter, and low-computation region, supporting the central premise that explicit interaction structure can reduce reliance on heavy stacking for deployable multi-entity reasoning.

By aligning architecture with problem structure, the proposed block turns our hypothesis into a direct empirical question: can explicit spatial, temporal, and cross spatio-temporal factorization provide the relational capacity that deep Transformer stacks usually learn implicitly? We answer this through the following contributions:
\begin{itemize}
    \item We propose a structure-first framework that factorizes multi-entity temporal dynamics into spatial, temporal, and cross spatio-temporal relational views, realized by a compact Transformer block with parallel self-attention, bidirectional cross-attention, and learnable gated fusion.

    \item We show that explicit interaction structure reduces reliance on depth: across depth studies, ablations, and multi-seed experiments, a single structured block achieves stable, competitive performance without deep Transformer stacking.

    \item We validate the framework across video-based group activity recognition, skeleton-based human--human interaction recognition, and wearable-sensor activity recognition, where entities correspond to persons, persons represented by skeleton sequences, and body-mounted sensor streams. Across modalities, the model achieves a strong accuracy - efficiency trade-off with about \(1.76\)M relational-block parameters, supporting edge-oriented and low-computation deployment.
\end{itemize}

\section{Related Work}

Multi-entity spatio-temporal modeling addresses problems where the target behavior emerges from multiple entities evolving together over time. In such settings, prediction depends not only on isolated appearance or motion cues, but also on inter-entity relations, temporal evolution, and how these relations change dynamically. Early CNN- and RNN-based models were effective for visual and temporal feature extraction, but offered limited mechanisms for explicit relational reasoning. This motivated a shift toward structure-aware models, first through graph-based formulations and later through attention-based architectures that learn interactions directly from data\cite{simonyan2014twostream, carreira2017quo, ibrahim2016hierarchical}.

\paragraph{Graph-based multi-entity spatio-temporal modeling: }
Graph-based methods represent entities as nodes and their relationships as edges, making relational structure explicit. This is naturally aligned with group activity recognition, human interaction understanding, and skeleton-based action recognition, where collective behavior depends on interactions among people, joints, or structured body representations. Actor-relation graphs model dependencies among people in group scenes, while spatial-temporal graph convolutional networks encode skeleton connectivity and temporal evolution through structured edges\cite{wu2019arg,yan2018stgcn,shi2019dgnn,chen2021ctrgcn}. These approaches established the importance of interaction-centric reasoning beyond isolated entity features\cite{pramono2020sacrf}. However, their effectiveness often depends on how the graph is defined, such as skeleton layouts, distance-based actor links, nearest-neighbor rules, or modality-specific connectivity\cite{chen2021ctrgcn, yuan2021din,ray2025wacv}. Such inductive bias is useful, but can become restrictive when interactions are dynamic, long-range, or vary across modalities, motivating more flexible attention-based alternatives\cite{yuan2021din,wang20233mformer,do2024skateformer}.

\paragraph{Transformer-based spatio-temporal reasoning: }
Transformer-based models replace fixed graph edges with data-dependent attention\cite{vaswani2017attention}, allowing relevant entities and time steps to be selected dynamically. This flexibility has made Transformers effective for video understanding\cite{wang2018nonlocal, bertasius2021timesformer}, group activity recognition\cite{gavrilyuk2020actortransformer, li2021groupformer, han2022dualai,chappa2023spartan}, and skeleton-based reasoning\cite{duan2022poseconv3d, wang20233mformer,do2024skateformer}. However, many spatio-temporal Transformers obtain relational capacity through deep stacks of generic attention and feed-forward layers, where spatial relations, temporal evolution, and their coupling are expected to emerge implicitly across depth~\cite{dosovitskiy2021vit, bertasius2021timesformer}. Even factorized designs that separate spatial and temporal attention often arrange them as repeated or sequential modules inside deeper architectures\cite{bertasius2021timesformer, gao2022fgstformer}. Similarly, cross-attention is commonly used for multimodal fusion, query-context aggregation, or late-stage information exchange\cite{chen2021crossvit}, rather than as an explicit mechanism for coupling spatial and temporal interaction views. As a result, the core dependencies of multi-entity temporal data are often mixed gradually through depth, increasing parameter cost, computation, and latency\cite{do2024skateformer, park2024calanet,zhang2025mopformer}.

Our work takes a structure-first view of Transformer design. Rather than treating attention as a generic token-mixing operation to be repeatedly stacked, we organize attention around the natural structure of multi-entity temporal data: spatial interactions among entities, temporal evolution of each entity, and cross spatio-temporal coupling between these views. Prior works have used spatial attention, temporal attention, cross-attention, and fusion mechanisms, but typically as separate components, task-specific modules, or parts of deeper pipelines\cite{bertasius2021timesformer, gavrilyuk2020actortransformer, li2021groupformer, han2022dualai, yuan2021din, chappa2023spartan}. In contrast, we integrate parallel spatial and temporal self-attention, bidirectional spatio-temporal cross-attention, and learnable gated fusion into a single compact relational block. This explicitly decomposes, couples, and fuses the essential interaction views, reducing reliance on depth while enabling an efficient framework that transfers across video, skeleton, and wearable-sensor domains.

\section{Methodology}

\begin{figure}[t]
    \centering
    \includegraphics[width=1.1\linewidth]{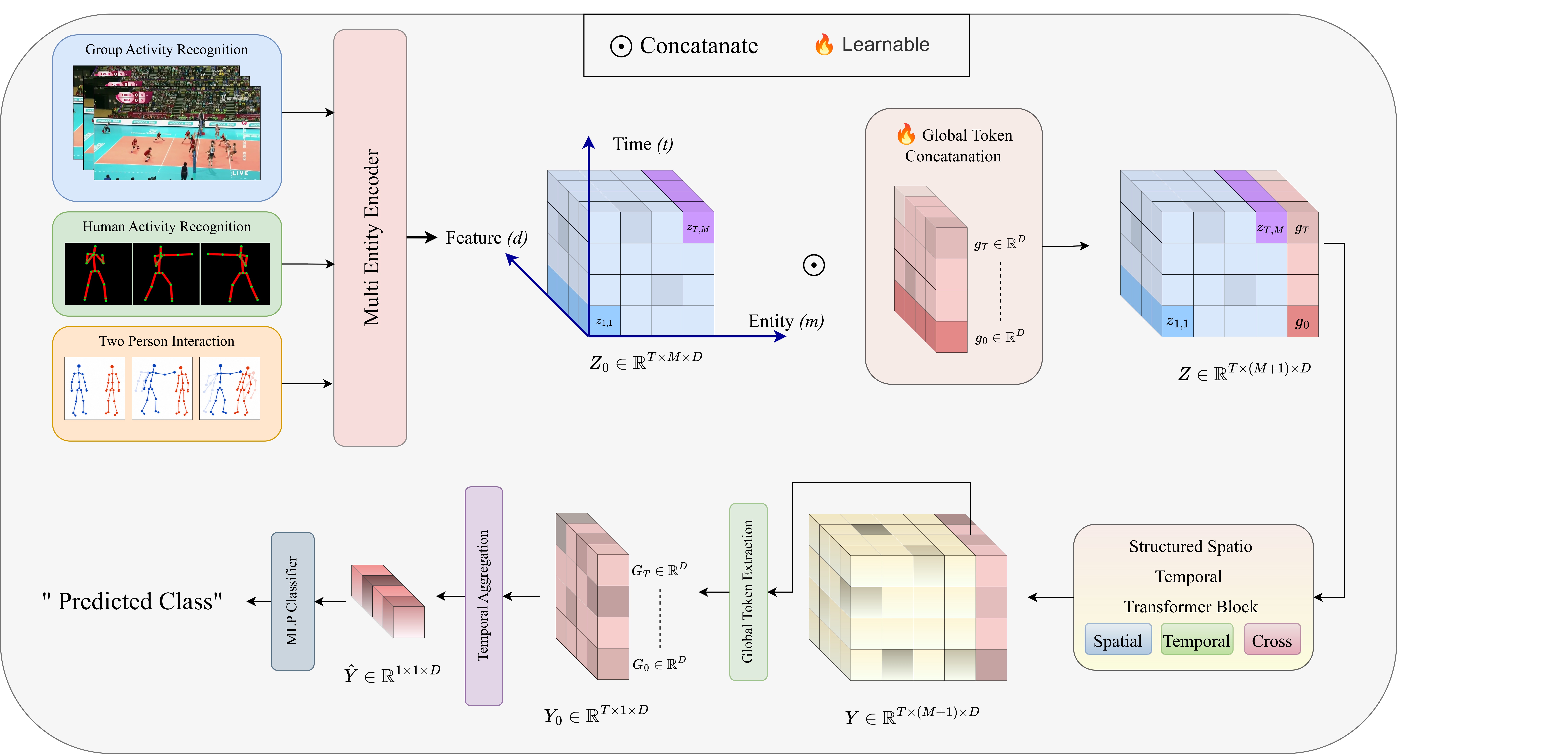}
    \caption{
    Overall pipeline of the proposed framework.
    }
    \label{fig:pipeline}
\end{figure}

As illustrated in Fig.~\ref{fig:pipeline}, we formulate multi-entity temporal recognition as a unified relational modeling problem. Given an input sequence $I$ from video, skeleton, or sensor modalities, a modality-specific encoder $\mathcal{E}_m$ maps the raw observations and available structural cues into a shared entity-time tensor
$Z_0 \in \mathbb{R}^{T \times M \times D}$, as shown in Fig.~\ref{fig:multidomain_pipeline}. We then append a learnable group token to each time step, producing
$Z \in \mathbb{R}^{T \times (M+1) \times D}$, which preserves entity-level evidence while providing a frame-level global state.

The augmented tensor is processed by the proposed structured spatio-temporal transformer block, temporally aggregated, and passed to a prediction head. The overall forward mapping is
\begin{equation}
\hat{y}
=
\psi
\left(
\mathcal{A}
\left(
\mathcal{B}_{\theta}
\left(
\mathcal{G}
\left(
\mathcal{E}_{m}(I)
\right)
\right)
\right)
\right),
\end{equation}
where $\mathcal{E}_{m}(I)=Z_0$ is the modality-specific encoder output, $\mathcal{G}$ denotes group-token augmentation, $\mathcal{B}_{\theta}$ is the structured spatio-temporal block, $\mathcal{A}$ denotes temporal aggregation, and $\psi$ is the prediction head.

The following subsections define each component of this pipeline in detail.

\subsection{Unified Multi-Entity Temporal Representation}

\begin{figure}[t]
    \centering
    \includegraphics[width=0.8\textwidth]{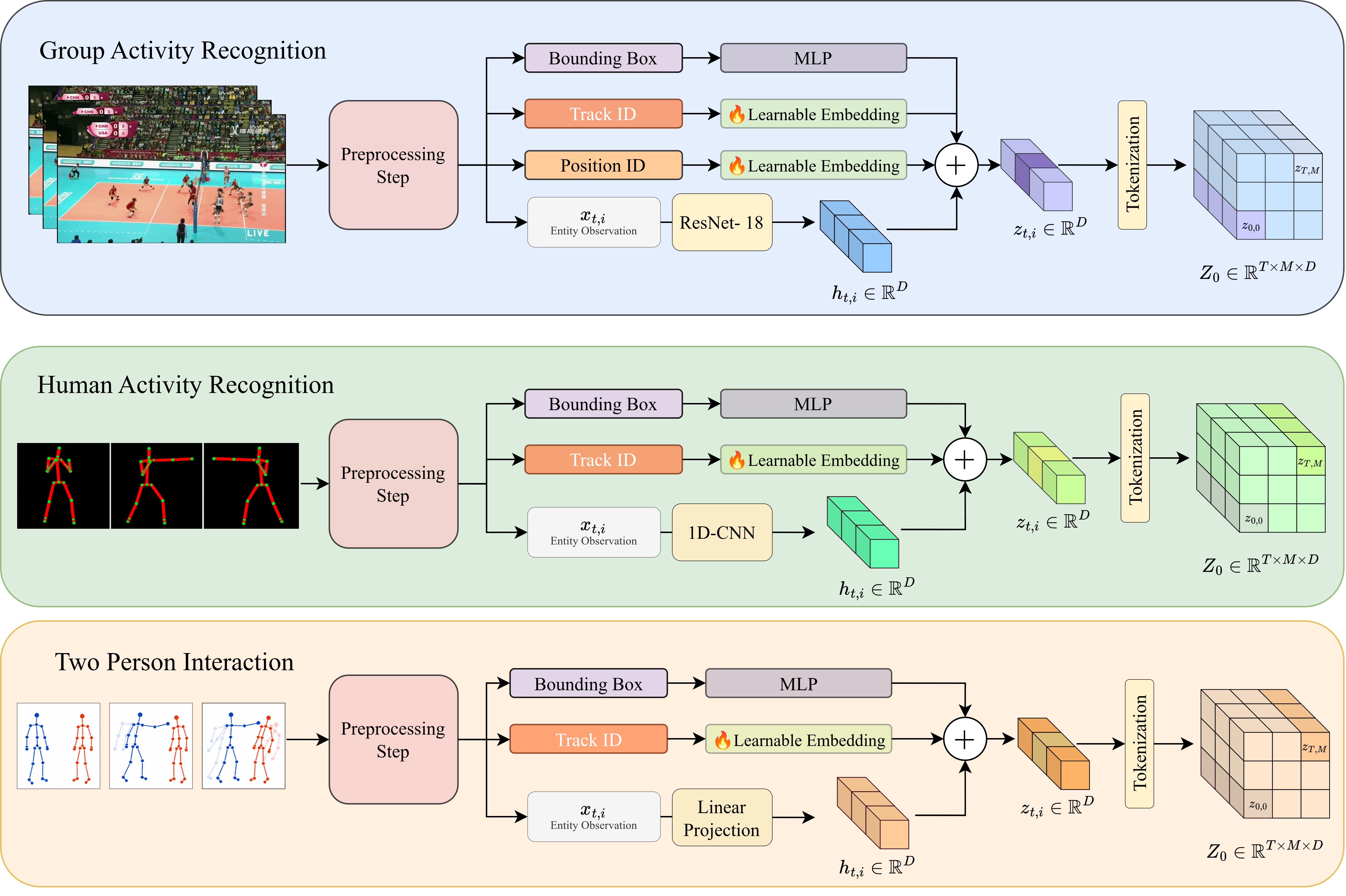}
    \caption[Unified Multi-Entity Temporal Representation]{
    Modality-specific entity standardization into a common tensor form, \(Z_0 \in \mathbb{R}^{T \times M \times D}\).
    }
    \label{fig:multidomain_pipeline}
\end{figure}

We first convert each raw input sequence $I$ into a unified entity-time tensor $Z_0$. The purpose of this stage is to remove modality-specific differences before relational reasoning: video crops, skeleton trajectories, and sensor streams have different raw formats, but they can all be represented as entities evolving over time.

As shown in Fig.~\ref{fig:multidomain_pipeline}, each input sequence is first passed through a task-specific preprocessing stage that extracts both entity observations and their available structural cues. These include the entity observation $x_{t,i}$, bounding-box geometry $b_{t,i}$, track identity $r_{t,i}$, and position identity $p_{t,i}$ when available. Mathematically this step can be represented as,
\begin{equation}
\mathcal{P}_{m}(I)
=
\left\{
x_{t,i}, b_{t,i}, r_{t,i}, p_{t,i}
\right\}_{t=1,i=1}^{T,M},
\end{equation}

Each entity observation ($x_{t,i}$) is then mapped into a shared $D$-dimensional feature space using a modality-specific encoder:
\begin{equation}
h_{t,i}=f^m_{\mathrm{enc}}(x_{t,i}) \in \mathbb{R}^{D}.
\end{equation}
Here, $f^m_{\mathrm{enc}}$ is implemented according to the input modality: a ResNet-18 encoder for GAR, a 1D convolutional encoder for HAR, and a Linear Projection for Two Person Interaction.

After obtaining the encoded entity feature $h_{t,i}$, we fuse it with the structural cues extracted during preprocessing. The bounding-box geometry $b_{t,i}$ is projected through an MLP to obtain $e^{\mathrm{box}}_{t,i}$, while the track identity $r_{t,i}$ and position identity $p_{t,i}$ index learnable embeddings $e^{\mathrm{id}}_{t,i}$ and $e^{\mathrm{pos}}_{t,i}$, respectively. The initial entity token is then constructed as
\begin{equation}
z_{t,i}
=
h_{t,i}
+
e^{\mathrm{box}}_{t,i}
+
e^{\mathrm{id}}_{t,i}
+
e^{\mathrm{pos}}_{t,i}.
\end{equation}

For modalities where a particular structural cue is unavailable, the corresponding embedding term is omitted. For example, the group activity branch uses bounding-box, track-ID, and position-ID embeddings, while the human activity and two-person interaction branches use the structural cues available from their respective preprocessing pipelines. This fusion allows each token to encode both the entity-level observation and its structural context before relational modeling.

The fused entity tokens are then passed through a tokenizer, where it stacks the tokens across all time steps and entity indices to form the unified multi-entity temporal representation:
\begin{equation}
Z_0 =
\mathrm{Tokenizer}\left(\{z_{t,i}\}_{t=1,i=1}^{T,M}\right)
\in \mathbb{R}^{T \times M \times D},
\end{equation}
where \(T\) is the number of time steps, \(M\) is the number of entities, and \(D\) is the token dimension. The tensor $Z_0$ is then passed to the group-token concatenation stage before structured spatio-temporal interaction modeling. This formulation $Z_0=\mathcal{E}_{m}(I)$ integrates appearance (what), position (where), temporal context (when), and entity identity (who) into a unified representation.

\subsection{Global Token Concatenation}

As shown in Fig.~\ref{fig:pipeline}, we append a learnable group token to each time step to provide a frame-level global representation. Given $Z_0 \in \mathbb{R}^{T \times M \times D}$, let $g \in \mathbb{R}^{D}$ be a learnable token denoted as $g_t$ at time step $t$. The group-token augmentation $\mathcal{G}$ produces,
\begin{equation}
Z
=
\left\{
[z_{t,1}, z_{t,2}, \ldots, z_{t,M}, g_t]
\right\}_{t=1}^{T},
\qquad
Z \in \mathbb{R}^{T \times (M+1) \times D}.
\end{equation}

The group token provides a compact frame-level state that can attend to all entities without prematurely pooling them. This preserves entity-level evidence while giving the transformer block a designated token for global context aggregation.



\begin{figure}[t]
    \centering
    \includegraphics[width=\linewidth]{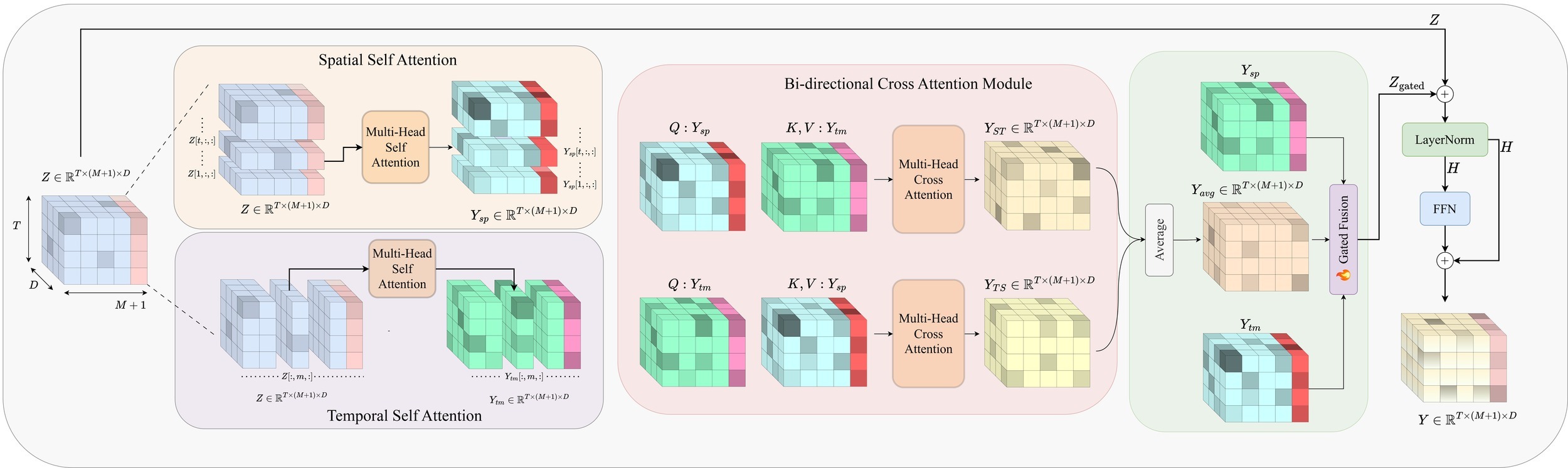}
    \caption{Our proposed structured spatio-temporal transformer block.}
    \label{fig:transformer_block}
\end{figure}

\subsection{Structured Spatio-Temporal Interaction Modeling}

We now model the interactions of the $Z$ using the proposed structured transformer block (Fig.~\ref{fig:transformer_block}). Instead of relying on deep stacked attention layers, we explicitly decompose interactions into three fundamental components: spatial, temporal, and cross-domain interactions. We denote multi-head attention~\cite{vaswani2017attention} as $\mathrm{MHA}(Q,K,V)$ in this section.

\paragraph{Spatial Interaction: }
As seen in Fig.\ref{fig:transformer_block}, spatial attention is applied independently at each time step over the entity dimension:

\begin{equation}
Y_{\mathrm{sp}}[t,:,:]
=
\mathrm{MHA}
\left(
Z[t,:,:],Z[t,:,:],Z[t,:,:]
\right).
\qquad
Y_{\mathrm{sp}} \in \mathbb{R}^{T \times (M+1) \times D}.
\end{equation}

This captures inter-entity dependencies such as spatial arrangement, relative positioning, and group structure.

\paragraph{Temporal Interaction: }
Temporal attention is applied independently to each entity slot across time. For a fixed entity or group-token index $i$, the temporal slice is $Z[:,i,:] \in \mathbb{R}^{T \times D}$. The temporal branch is computed as
\begin{equation}
Y_{\mathrm{tm}}[:,i,:]
=
\mathrm{MHA}
\left(
Z[:,i,:],Z[:,i,:],Z[:,i,:]
\right),
\qquad
Y_{\mathrm{tm}} \in \mathbb{R}^{T \times (M+1) \times D}.
\end{equation}
This models the evolution of each entity token over time without mixing different entity slots inside the temporal branch. Thus, spatial attention captures relations within each frame, while temporal attention captures how each token evolves across frames.

\paragraph{Cross Interaction: }
While spatial and temporal attention capture intra-domain dependencies, they do not directly connect space and time. To bridge this gap, we introduce bidirectional cross-attention:
\begin{equation}
Y_{\text{avg}} = \frac{1}{2} \left(
\text{MHA}(Y_{\text{sp}}, Y_{\text{tm}}, Y_{\text{tm}}) +
\text{MHA}(Y_{\text{tm}}, Y_{\text{sp}}, Y_{\text{sp}})
\right),
\end{equation}
where, the first direction updates spatial features using temporal context, while the second updates temporal features using spatial context. The final cross-interaction representation $Y_{avg}$ is obtained by averaging the two directions.

\paragraph{Gated Fusion: }
The three interaction streams are combined using a learnable gating mechanism:
\begin{equation}
Z_{\text{gated}} = g_{\text{sp}} Y_{\text{sp}} + g_{\text{tm}} Y_{\text{tm}} + g_{\text{avg}} Y_{\text{avg}}, 
\quad g_{\text{sp}} + g_{\text{tm}} + g_{\text{avg}} = 1
\end{equation}

The gating weights provide an interpretable mechanism that adaptively selects the importance of each interaction type for a given input.

The fused update $Z_{\text{gated}}$ is added to the input $Z$ through a residual connection and normalized:
\begin{equation}
H
=
\mathrm{LayerNorm}
\left(
Z + Z_{\mathrm{gated}}
\right).
\end{equation}
The final block output is obtained using a position-wise feed-forward network followed by a second residual normalization:
\begin{equation}
Y
=
\mathrm{LayerNorm}
\left(
H+\mathrm{FFN}(H)
\right),
\qquad
Y \in \mathbb{R}^{T \times (M+1) \times D}.
\end{equation}

\subsection{Temporal Aggregation and Prediction }

As represented in Fig. \ref{fig:pipeline}, from the output of the Structured Spatio-Temporal Transformer Block $Y$, we extract group tokens across time:
\begin{equation}
Y_0 = [G_1, \dots, G_T]
\end{equation}

where $G_t$ is the contextualized group-token representation at time step $t$.

These are aggregated using attention pooling:
\begin{equation}
h = \sum_{t=1}^{T} \alpha_t G_t, \quad
\alpha_t = \frac{\exp(w^T G_t)}{\sum_k \exp(w^T G_k)}
\end{equation}

Finally, the aggregated representation is passed to a classifier:
\begin{equation}
\hat{y} = \text{MLP}(h)
\end{equation}

This final stage converts the structured relational representation into a compact sequence descriptor for prediction. 

Overall, the proposed architecture is motivated by the view that multi-entity temporal reasoning can be factorized into three complementary interaction views: spatial relations among entities, temporal evolution of entity states, and cross spatio-temporal coupling. By explicitly exposing these views within a single structured relational block, the model reduces the burden on deep generic attention stacks to discover them implicitly, yielding an efficient and interpretable architecture for multi-entity temporal recognition.

\section{Experiments and Results}

Our experiments test the central claim of this work: explicit spatial, temporal, and cross spatio-temporal interaction modeling can provide strong multi-entity reasoning with reduced reliance on deep Transformer stacking. We evaluate this claim along four axes: accuracy, efficiency, component necessity, and depth sensitivity.

\paragraph{Experimental setup.}
We evaluate the same structured relational block across three domains: video group activity recognition, wearable sensor-based HAR, and skeleton-based two-person interaction recognition. We use Volleyball and Collective Activity for video, PAMAP2, Opportunity, and UCI-HAR for sensors, and the mutual-interaction subsets of NTU RGB+D 60/120 for skeleton interactions. Across domains, only the modality-specific entity standardization front-end changes: ResNet-18 for video, a lightweight 1D CNN for sensor streams, and a lightweight projection encoder for skeleton features. For sensor inputs, we use standard sliding-window preprocessing, with overlapping windows generated only within fixed subject/session/run splits; normalization statistics are computed from the training split only to prevent leakage between temporally adjacent windows. Unless otherwise stated, all experiments use one structured block with \(D=256\) and 8 attention heads and we train the proposed model using the standard cross-entropy loss. We report accuracy and macro F1 over multiple seeds, with no test-time augmentation, temporal ensembling, or model ensembling. Full protocols, preprocessing, hyperparameters, FLOP-counting details, mean--standard deviation results, and code release details are provided in the appendix.

\paragraph{Accuracy - efficiency trade-off:}
Tables~\ref{tab:volleyball_collective_acc}, \ref{tab:har_comparison}, and \ref{tab:ntu_interaction_comparison} show that the proposed framework consistently achieves a strong accuracy--efficiency trade-off across video, sensor, and skeleton domains. On group activity recognition, it matches or surpasses strong prior methods on Volleyball and Collective Activity while using a lightweight ResNet-18 backbone and substantially fewer FLOPs than several deeper alternatives. On sensor-based HAR, the same relational block obtains the best accuracy and macro F1 on PAMAP2, Opportunity, and UCI-HAR, showing that the formulation transfers from visual actors to sensor-stream entities. On skeleton-based two-person interaction recognition, it remains competitive with recent graph- and Transformer-based methods on NTU RGB+D 60/120 while using only about \(1.76\)M parameters and \(0.49\) GFLOPs. These results indicate that explicit interaction structure can preserve strong recognition performance while reducing the computational burden of deep spatio-temporal stacking.

\begin{figure}[t]
    \centering

    \begin{subfigure}{0.32\linewidth}
        \centering
        \includegraphics[width=\linewidth]{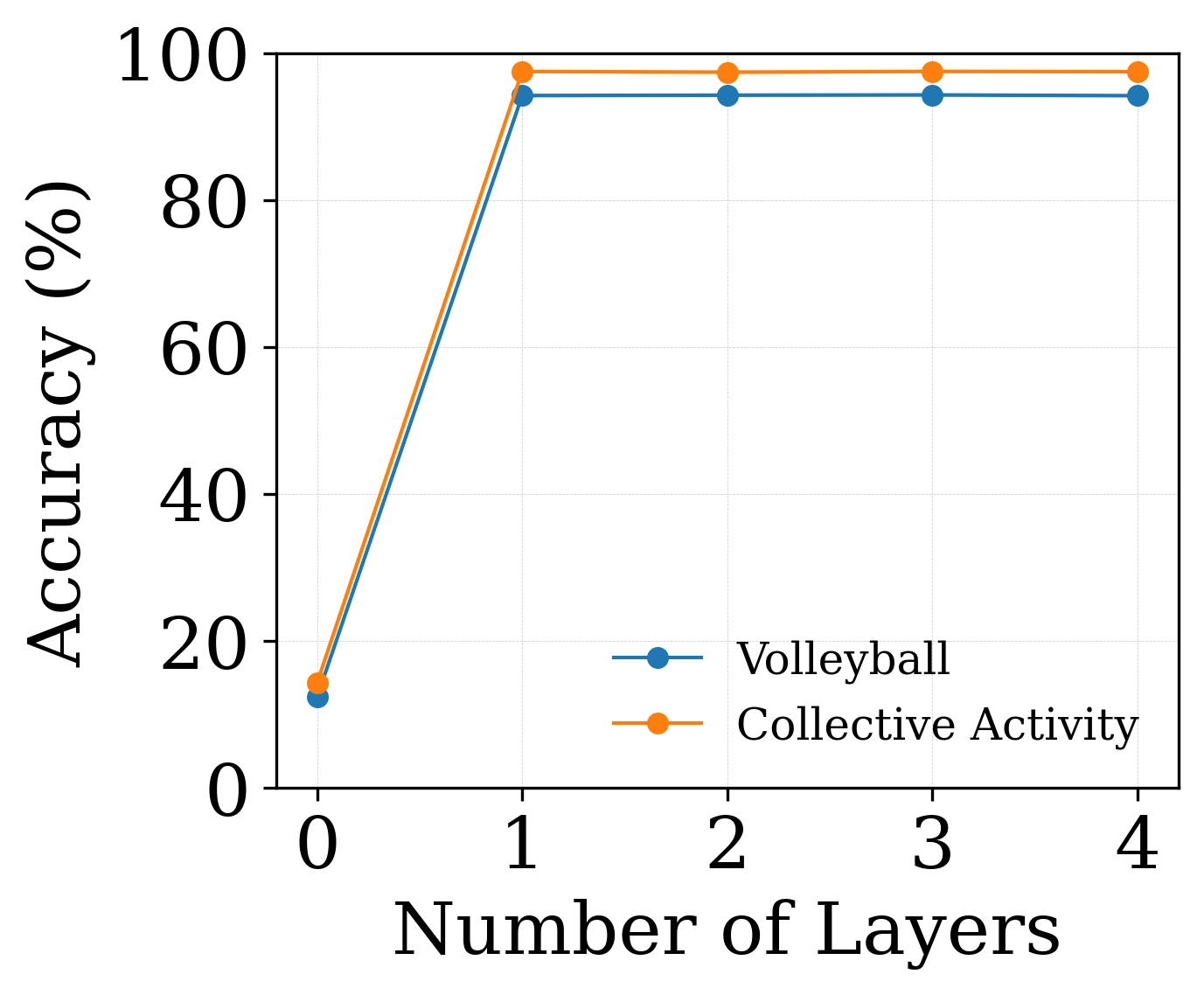}
        \caption{Group activity.}
        \label{fig:layers_group_activity}
    \end{subfigure}
    \hfill
    \begin{subfigure}{0.32\linewidth}
        \centering
        \includegraphics[width=\linewidth]{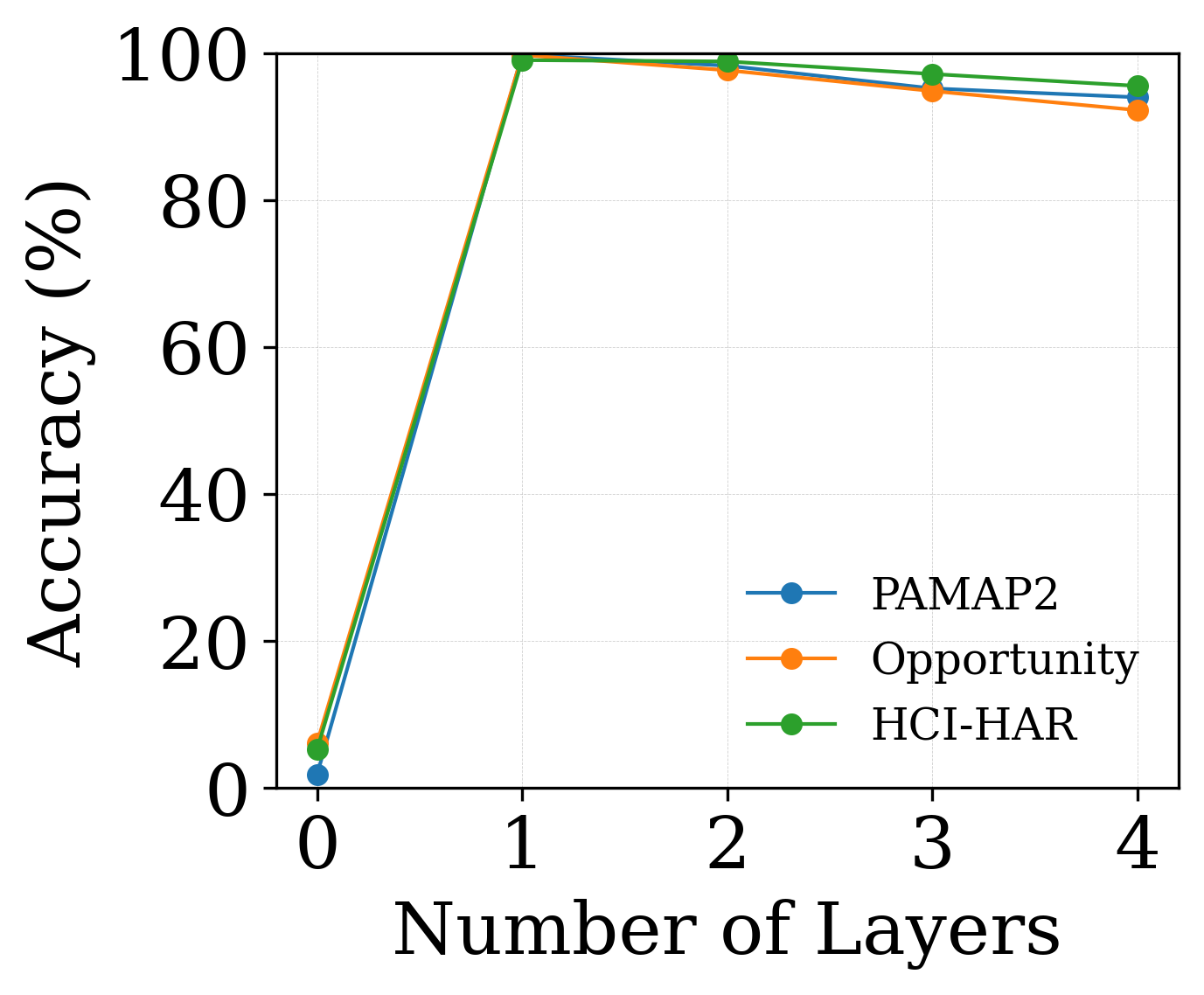}
        \caption{Human action recognition.}
        \label{fig:layers_har}
    \end{subfigure}
    \hfill
    \begin{subfigure}{0.32\linewidth}
        \centering
        \includegraphics[width=\linewidth]{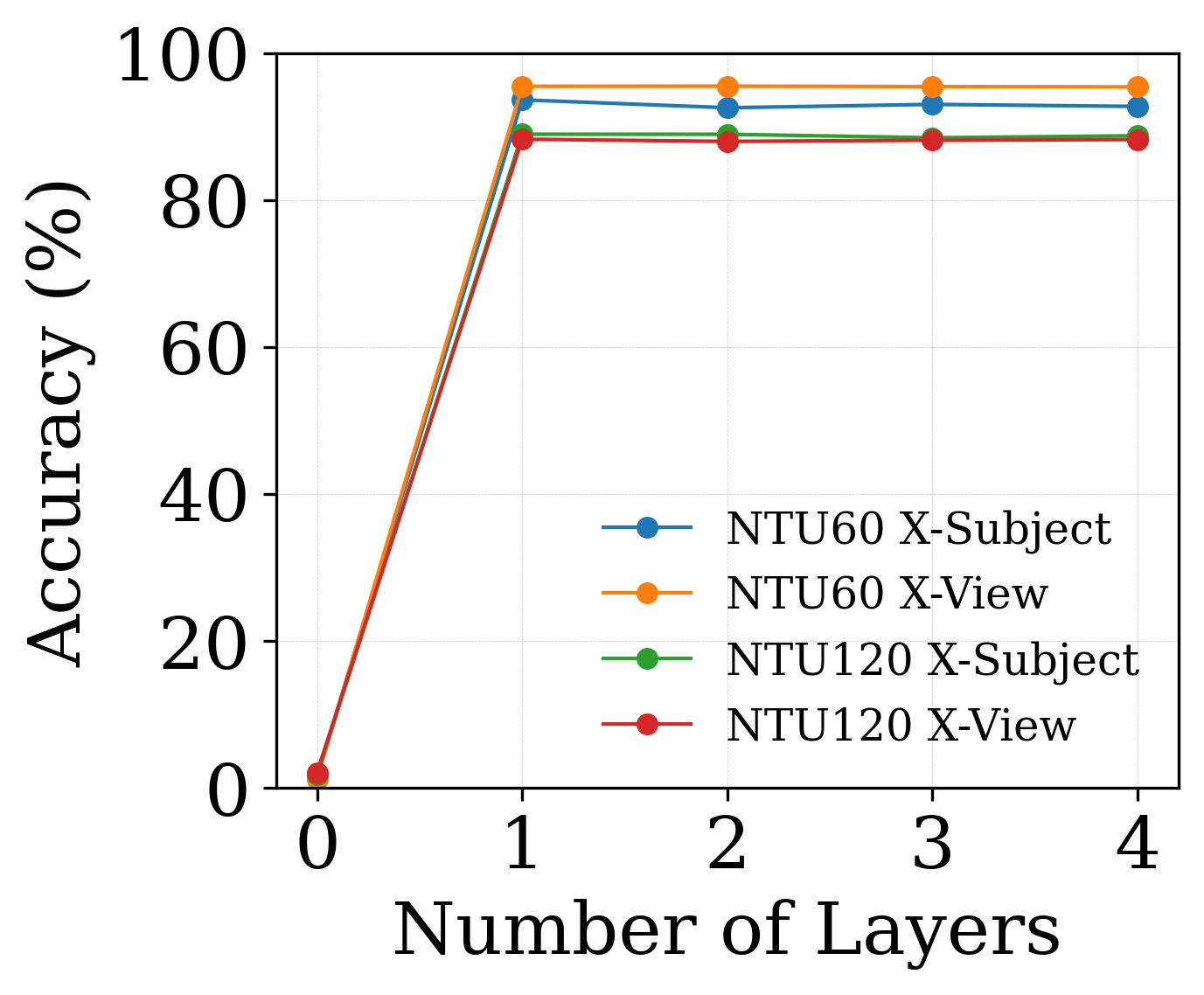}
        \caption{Two-person interaction.}
        \label{fig:layers_two_person}
    \end{subfigure}

    \caption{
    Depth of our blocks VS Performance
    }
    \label{fig:layer_sensitivity_all}
\end{figure}

\paragraph{Depth sensitivity.}
Fig.~\ref{fig:layer_sensitivity_all} evaluates the effect of stacking additional structured blocks. Across video, sensor, and skeleton modalities, the dominant improvement comes from the first block, while deeper stacking provides limited or inconsistent gains and can slightly degrade performance. This supports the structure-over-depth hypothesis: explicitly modeling spatial, temporal, and cross spatio-temporal interactions within a single stage reduces the need for repeated stacking while preserving efficiency.

\begin{table}[t]
\centering
\caption{
Comparison with previous methods on the Volleyball and Collective Activity datasets.
Red denotes the best result, blue denotes the second-best result, and bold denotes the third-best result in each metric column.
For accuracy metrics, higher is better; for Params and GFLOPs, lower is better.
}
\label{tab:volleyball_collective_acc}
\footnotesize
\setlength{\tabcolsep}{4pt}
\renewcommand{\arraystretch}{1.08}
\resizebox{\linewidth}{!}{%
\begin{tabular}{@{}lccccccc@{}}
\toprule
\textbf{Method}
& \multicolumn{5}{c}{\textbf{Volleyball Dataset}}
& \multicolumn{2}{c}{\textbf{Collective Activity Dataset}} \\
\cmidrule(lr){2-6} \cmidrule(lr){7-8}
& \textbf{Backbone}
& \textbf{MCA}
& \textbf{Merged MCA}
& \textbf{Params}
& \textbf{GFLOPs}
& \textbf{Backbone}
& \textbf{MPCA} \\
\midrule
PCTDM\cite{yan2018pctdm}         
& ResNet-18    
& 90.3 
& 94.3 
& --    
& --    
& AlexNet      
& 92.2  \\

StagNet\cite{qi2018stagnet}       
& VGG-16       
& 89.3 
& --   
& --    
& --    
& VGG-16       
& 89.1  \\

ARG\cite{wu2019arg}           
& ResNet-18    
& 91.1 
& \textbf{95.1} 
& \textbf{49.5M} 
& \textbf{307G}  
& ResNet-18    
& 92.3  \\

CRM\cite{azar2019crm}           
& I3D          
& 92.1 
& --   
& --    
& --    
& I3D          
& 94.2  \\

HiGCIN\cite{yan2020higcin}        
& ResNet-18    
& 91.4 
& --   
& --    
& --    
& ResNet-18    
& 93.0  \\

AT\cite{gavrilyuk2020actortransformer}            
& ResNet-18    
& 90.0 
& 94.0 
& --    
& --    
& --           
& --    \\

SACRF\cite{pramono2020sacrf}         
& ResNet-18    
& 90.7 
& 92.7 
& 53.7M 
& 339G  
& --           
& --    \\

DIN\cite{yuan2021din}           
& ResNet-18    
& 93.1 
& \textcolor{red}{95.6} 
& \textcolor{blue}{26.0M} 
& \textcolor{blue}{304G}  
& ResNet-18    
& \textbf{95.3}  \\

TCE+STBiP\cite{yuan2021tce}     
& VGG-16       
& \textbf{94.1} 
& --   
& --    
& --    
& Inception-v3 
& 95.1  \\

GroupFormer\cite{li2021groupformer}   
& Inception-v3 
& \textbf{94.1} 
& --   
& --    
& --    
& --           
& --    \\

Dual-AI (RGB)\cite{han2022dualai} 
& Inception-v3 
& \textcolor{blue}{94.4} 
& --   
& --    
& --    
& Inception-v3 
& \textcolor{blue}{96.5}  \\

\midrule
\textbf{Ours} 
& \textbf{ResNet-18} 
& \textcolor{red}{94.6} 
& \textcolor{blue}{95.3} 
& \textcolor{red}{13.38M} 
& \textcolor{red}{18G}
& \textbf{ResNet-18} 
& \textcolor{red}{97.47} \\
\bottomrule
\end{tabular}%
}
\end{table}

\captionof{table}{
Comparison of accuracy and F1 score on PAMAP2, Opportunity, and UCI-HAR datasets.
Red denotes the best result, blue denotes the second-best result, and bold denotes the third-best result in each column.
}
\label{tab:har_comparison}
\setlength{\tabcolsep}{6pt}
\renewcommand{\arraystretch}{1.15}
\resizebox{\linewidth}{!}{%
\begin{tabular}{@{}lcccccc@{}}
\toprule
\textbf{Method}
& \multicolumn{2}{c}{\textbf{PAMAP2}}
& \multicolumn{2}{c}{\textbf{Opportunity}}
& \multicolumn{2}{c}{\textbf{UCI-HAR}} \\
\cmidrule(lr){2-3} \cmidrule(lr){4-5} \cmidrule(lr){6-7}
& \textbf{Accuracy (\%)}
& \textbf{F1 Score (\%)}
& \textbf{Accuracy (\%)}
& \textbf{F1 Score (\%)}
& \textbf{Accuracy (\%)}
& \textbf{F1 Score (\%)} \\
\midrule
Layer-wise CNN\cite{teng2020layerwisecnn} 
& 92.97 & 93.03 & 81.00 & 80.55 & 96.98 & 96.97 \\

LSTM-CNN\cite{xia2020lstmcnn}       
& -- & -- & 92.63 & 92.63 & 95.78 & 95.18 \\

Attn-HAR\cite{mahmud2020attnhar}       
& 96.00 & 96.00 & 67.00 & 42.00 & -- & -- \\

DanHAR\cite{gao2021danhar}         
& 93.16 & -- & 82.75 & -- & -- & -- \\

Contrast-HAR\cite{cheng2023contrasthar}   
& 93.22 & 92.97 & -- & -- & 98.00 & 98.00 \\

AFVF\cite{nguyen2024afvf}           
& \textbf{96.72} & \textbf{96.65} & -- & -- & \textbf{98.61} & \textbf{98.65} \\

MLCNNwav\cite{dahou2024mlcnnwav}       
& -- & -- & \textbf{93.19} & \textbf{93.30} & 95.52 & 96.11 \\

STDual-X\cite{chandirakumar2025stdualx}      
& \textcolor{blue}{97.02} 
& \textcolor{blue}{97.04} 
& \textcolor{blue}{95.43} 
& \textcolor{blue}{94.99} 
& \textcolor{blue}{98.80} 
& \textcolor{blue}{98.87} \\

\midrule
\textbf{Ours}  
& \textcolor{red}{99.76} 
& \textcolor{red}{99.68} 
& \textcolor{red}{99.71} 
& \textcolor{red}{99.58} 
& \textcolor{red}{99.02} 
& \textcolor{red}{99.14} \\
\bottomrule
\end{tabular}%
}


\captionof{table}{
Comparison with previous methods on the interaction subsets of NTU RGB+D 60 and NTU RGB+D 120. 
Red denotes the best result, blue denotes the second-best result, and bold denotes the third-best result in each column. 
For accuracy columns, higher is better; for Params and FLOPs, lower is better.
}
\label{tab:ntu_interaction_comparison}
\setlength{\tabcolsep}{4pt}
\renewcommand{\arraystretch}{1.08}
\resizebox{\linewidth}{!}{%
\begin{tabular}{@{}lcccccc@{}}
\toprule
\textbf{Method}
& \multicolumn{2}{c}{\textbf{NTU RGB+D 60}}
& \multicolumn{2}{c}{\textbf{NTU RGB+D 120}}
& \textbf{Param.(M)}
& \textbf{FLOPs(G)} \\
\cmidrule(lr){2-3} \cmidrule(lr){4-5}
& \textbf{Cross-subject (\%)}
& \textbf{Cross-view (\%)}
& \textbf{Cross-subject (\%)}
& \textbf{Cross-set (\%)}
&  &  \\
\midrule
SGN\cite{zhang2020sgn}          
& 89.0 & 94.5 & 79.2 & 81.5 & \textcolor{blue}{0.62} & -- \\

3s RA-GCN\cite{song2021ragcn}    
& 87.3 & 93.6 & 81.1 & 82.7 & 6.21 & -- \\

LSTM-IRN\cite{perez2022irn}     
& 90.5 & 93.5 & 77.7 & 79.6 & 5.76 & -- \\

Lee et al.\cite{lee2022interacting}   
& 89.6 & 90.7 & -- & -- & -- & -- \\

CoAGCN\cite{hedegaard2023continual}       
& 84.1 & 92.6 & 84.0 & 85.3 & 3.47 & \textcolor{blue}{0.17} \\

CoS-TR\cite{hedegaard2023continual}       
& 86.3 & 92.4 & 84.8 & 86.1 & 3.09 & \textcolor{red}{0.15} \\

3S-AimCLR++\cite{guo2024aimclr}  
& 87.1 & 93.0 & 82.5 & 83.2 & -- & -- \\

AS-GCN\cite{li2019asgcn}       
& 89.3 & 93.0 & 82.9 & 83.7 & 6.99 & -- \\

ST-GCN\cite{yan2018stgcn}       
& 83.3 & 88.7 & 80.2 & 79.0 & 3.14 & 16.73 \\

2s-AGCN\cite{shi2019agcn}      
& -- & -- & 86.1 & 88.1 & 6.94 & -- \\

ISTA-Net\cite{wen2023istanet}     
& -- & -- & \textcolor{red}{90.5} & \textcolor{red}{91.7} & 6.22 & 68.18 \\

IGFormer\cite{pang2022igformer}     
& \textbf{93.6} & \textcolor{blue}{96.5} & 86.5 & 85.4 & -- & -- \\

SkeleTR\cite{duan2023skeletr}      
& \textcolor{red}{94.9} & \textcolor{red}{97.7} & 88.3 & 87.8 & 3.82 & 7.30 \\

2D-DWT\cite{wu2025skeleton2ddwt}       
& \textbf{93.6} & 94.6 & \textbf{88.7} & \textcolor{blue}{89.3} & \textcolor{red}{0.53} & 1.89 \\

\midrule
\textbf{Ours} 
& \textcolor{blue}{93.62} 
& \textbf{95.46} 
& \textcolor{blue}{88.94} 
& \textbf{88.26} 
& \textbf{1.76} 
& \textbf{0.49} \\
\bottomrule
\end{tabular}%
}

\begin{figure}[t]
    \centering
    \includegraphics[
        width=\textwidth,
        height=0.25\textheight,
        keepaspectratio
    ]{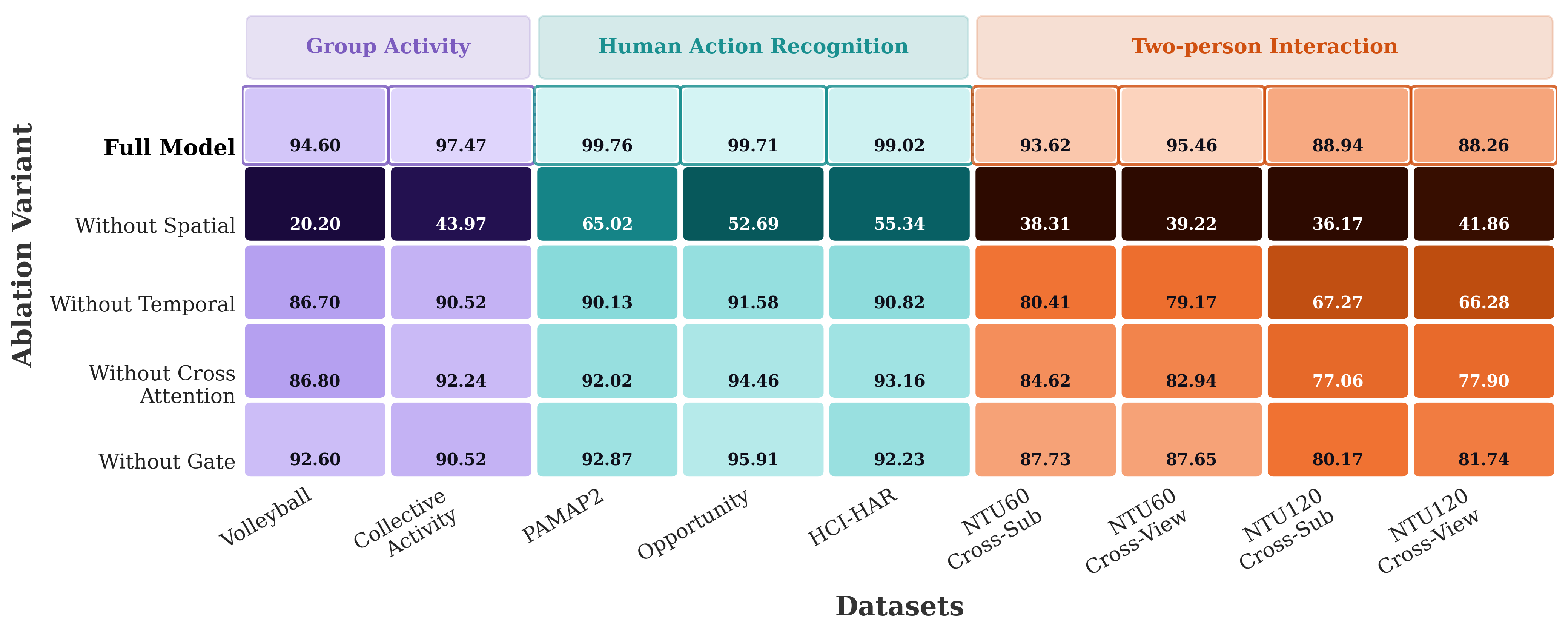}
    \caption{
    Component ablation study
    }
    \label{fig:ablation_heatmap}
\end{figure}

\paragraph{Component ablation.}
Fig.~\ref{fig:ablation_heatmap} evaluates the contribution of each interaction branch. Removing spatial, temporal, or cross-interaction modeling consistently reduces performance across task families, showing that the three views provide complementary relational information. Spatial ablation causes the largest degradation, indicating that spatial relations are particularly important for collective spatio-temporal reasoning. These results support explicitly decomposing and fusing interaction types instead of relying on generic attention alone.

\section{Conclusion}

This work shows that effective multi-entity spatio-temporal reasoning does not necessarily require deep Transformer stacking always. By explicitly factorizing interactions into spatial, temporal, and cross spatio-temporal views, the proposed single structured relational block achieves a strong accuracy--efficiency trade-off across video, skeleton, and wearable-sensor domains. The framework reaches competitive or superior performance on group activity recognition while reducing computation by about \(17\times\)--\(19\times\) compared with several prior high-performing models. On skeleton-based interaction recognition, it remains competitive with recent graph- and Transformer-based approaches using only \(1.76\)M parameters and \(0.49\) GFLOPs. On sensor-based HAR, the same relational block achieves the best reported accuracy and macro F1 across the evaluated benchmarks, demonstrating transferability beyond vision to signal-based entities. Together with the depth and ablation studies, these results provide empirical evidence that explicit interaction structure can reduce reliance on depth, enabling accurate, lightweight, and deployable multi-entity reasoning for edge-oriented and real-time applications.

\section{Limitation}
This work provides empirical evidence that explicit interaction decomposition can reduce reliance on deep Transformer stacking across multiple domains. However, a formal theoretical analysis of this behavior remains future work. In addition, performance may depend on the quality of modality-specific entity representations and preprocessing pipelines.

\bibliographystyle{unsrt}
\bibliography{references}

\newpage
\section{Appendix}

\subsection{Datasets}
\label{app:datasets}

We evaluate the proposed model across three families of multi-entity temporal recognition tasks: skeleton-based interaction recognition, video-based group activity recognition, and sensor-based human activity recognition. This evaluation is intended to test whether the proposed structured spatio-temporal block generalizes beyond a single input modality.

\paragraph{Skeleton-Based Interaction Recognition.}
We evaluate on the mutual-interaction subsets of NTU RGB+D 60 and NTU RGB+D 120. NTU RGB+D 60 provides 3D skeleton sequences with 25 joints per person and includes 11 two-person interaction classes. NTU RGB+D 120 extends this setting with additional action categories and includes 26 mutual-interaction classes in total. We follow the standard cross-subject and cross-view protocols for NTU RGB+D 60, and the cross-subject and cross-setup protocols for NTU RGB+D 120.

\paragraph{Group Activity Recognition.}
We evaluate on the Volleyball dataset and the Collective Activity Dataset (CAD). The Volleyball dataset contains group activity labels, individual action labels, and player bounding boxes. The task is to classify the group activity of each clip from multiple interacting players. CAD contains multi-person video sequences with person-level annotations and group-level activity labels. We follow the commonly used evaluation protocol for each dataset and report group-level classification performance.

\paragraph{Sensor-Based Human Activity Recognition.}
We evaluate on PAMAP2, Opportunity, and UCI-HAR. These datasets contain wearable or smartphone sensor signals collected from human activities. We treat sensor locations as entities and temporal sub-windows as time steps, allowing the same structured spatio-temporal block to be applied without modality-specific architectural changes.

\subsection{Evaluation Protocols}
\label{app:evaluation_protocols}

\paragraph{Skeleton-Based Interaction Recognition.}
For NTU RGB+D 60, we report top-1 accuracy under the cross-subject and cross-view protocols. For NTU RGB+D 120, we report top-1 accuracy under the cross-subject and cross-setup protocols. These protocols test whether the model generalizes across unseen subjects, camera views, or capture setups.

\paragraph{Group Activity Recognition.}
For Volleyball and CAD, we report group-level classification accuracy. For Volleyball, we additionally report merged-class accuracy when comparing with prior methods that merge semantically related classes. For CAD, we report the standard group activity classification accuracy used in prior work.

\paragraph{Sensor-Based Human Activity Recognition.}
For PAMAP2, Opportunity, and UCI-HAR, we report accuracy and macro F1 score. Macro F1 is included because sensor-based activity datasets can be class-imbalanced, and accuracy alone may overestimate performance on frequent classes.

\begin{table}[h]
\centering
\caption{Evaluation protocols and metrics used for each benchmark.}
\label{tab:eval_protocols}
\small
\setlength{\tabcolsep}{5pt}
\renewcommand{\arraystretch}{1.08}
\begin{tabular}{@{}llll@{}}
\toprule
\textbf{Task} 
& \textbf{Dataset} 
& \textbf{Protocol} 
& \textbf{Metric} \\
\midrule
Skeleton action recognition 
& NTU RGB+D 60 
& Cross-subject 
& Top-1 accuracy \\

Skeleton action recognition 
& NTU RGB+D 60 
& Cross-view 
& Top-1 accuracy \\

Skeleton action recognition 
& NTU RGB+D 120 
& Cross-subject 
& Top-1 accuracy \\

Skeleton action recognition 
& NTU RGB+D 120 
& Cross-View 
& Top-1 accuracy \\

\midrule
Group activity recognition 
& Volleyball 
& Standard split 
& MCA, MMCA \\

Group activity recognition 
& Collective Activity 
& Standard split 
& MPCA \\

\midrule
Sensor-based HAR 
& PAMAP2 
& Standard split 
& Accuracy, F1 score \\

Sensor-based HAR 
& Opportunity 
& Standard split 
& Accuracy, F1 score \\

Sensor-based HAR 
& UCI-HAR 
& Standard split 
& Accuracy, F1 score \\
\bottomrule
\end{tabular}
\end{table}

\subsection{Implementation Details}
\label{app:implementation}

\paragraph{Training.}
All models are trained end-to-end using AdamW with gradient clipping at a global norm of 1.0. We use dataset-specific learning-rate schedules following the convergence behavior of each task. Video and sensor experiments use step decay schedules, while skeleton experiments use cosine annealing. The exact hyperparameters are reported in Table~\ref{tab:full_hyperparams}.

\paragraph{Regularization.}
We use dropout and weight decay according to the dataset configuration. For skeleton datasets, we additionally use label smoothing to reduce over-confident predictions. For NTU RGB+D 120, focal loss is used to improve robustness under class imbalance and visually similar interaction categories.

\subsection{Model Configuration}
\label{app:model_config}

Table~\ref{tab:full_hyperparams} summarizes the architecture and training hyperparameters used for each dataset. The same structured spatio-temporal block is used across all modalities. Only the input encoder, token dimension, depth, dropout, and training schedule are adjusted for each dataset.

\begin{table}[h]
  \centering
  \small
  \caption{Complete hyperparameter configuration across datasets. $L$ denotes the number of structured blocks, $D$ the token dimension, $H$ the number of attention heads, $d_{\mathrm{ff}}$ the feed-forward hidden dimension, WD the weight decay, BS the batch size, and $E$ the number of training epochs.}
  \label{tab:full_hyperparams}
  \resizebox{\linewidth}{!}{%
  \begin{tabular}{lcccccccccc}
    \toprule
    \textbf{Dataset} & $L$ & $D$ & $H$ & $d_{\mathrm{ff}}$ & Drop. & LR & WD & BS & $E$ & Encoder \\
    \midrule
    Volleyball  & 1 & 256 & 8 & 1024 & 0.1 & $10^{-4}$ & $10^{-4}$ & 4  & 100  & ResNet-18 \\
    CAD         & 1 & 256 & 8 & 512  & 0.3 & $3{\times}10^{-4}$ & $10^{-4}$ & 8  & 100  & ResNet-18 \\
    NTU60       & 1 & 256 & 8 & 512  & 0.2 & $10^{-3}$ & $5{\times}10^{-4}$ & 64 & 500 & Linear Projection encoder \\
    NTU120      & 1 & 256 & 8 & 512  & 0.3 & $3{\times}10^{-4}$ & $10^{-3}$ & 64 & 500 & Linear Projection encoder \\
    PAMAP2      & 1 & 256 & 8 & 1024 & 0.3 & $10^{-3}$ & $10^{-4}$ & 64 & 200 & 1D CNN \\
    Opportunity & 1 & 256 & 8 & 1024 & 0.3 & $10^{-3}$ & $10^{-4}$ & 64 & 200 & 1D CNN \\
    UCI-HAR     & 1 & 256 & 8 & 1024 & 0.4 & $10^{-3}$ & $5{\times}10^{-4}$ & 64 & 200 & 1D CNN \\
    \bottomrule
  \end{tabular}}
\end{table}

\subsection{Input Representation}
\label{app:input_repr}

Each input sequence is represented (after modality-specific encoding) as an entity--time tensor
\begin{equation}
Z_0 \in \mathbb{R}^{T \times M \times D},
\end{equation}
where $T$ is the number of time steps, $M$ is the number of entities, and $D$ is the token dimension.

\paragraph{Video Inputs.}
For Volleyball and CAD, each person crop is encoded using a ResNet-18 backbone pretrained on ImageNet. The final classification layer is removed, and the pooled feature vector is projected to the model dimension. Bounding-box coordinates are encoded using a learned spatial projection and added to the visual token. Temporal and identity embeddings are also added to preserve frame order and entity consistency.

\paragraph{Skeleton Inputs.}
For NTU RGB+D 60 and NTU RGB+D 120, each person is treated as an entity. For NTU RGB+D 60, we construct a 300-dimensional skeleton descriptor using four components: body-centered joint coordinates, absolute joint coordinates, inter-person relative coordinates, and temporal velocity. For NTU RGB+D 120, we additionally include bone vectors and bone velocity, resulting in a 444-dimensional descriptor. These descriptors are projected to the model dimension using a linear layer.

\paragraph{Sensor Inputs.}
For PAMAP2, Opportunity, and UCI-HAR, sensor locations are treated as entities. Each sensor stream is divided into temporal sub-windows and encoded using a lightweight 1D CNN. The resulting sub-window features are projected to the shared token dimension and processed by the same structured spatio-temporal block.

\subsection{Positional and Identity Encodings}
\label{app:encodings}

We use three types of encodings depending on the dataset: spatial encoding, temporal encoding, and identity encoding. Spatial encodings represent entity location or sensor position. Temporal encodings represent the time index. Identity encodings preserve entity identity across time, such as player ID, person slot, or sensor location. These encodings are summed with the entity token before the structured block.

\begin{figure*}[t]
    \centering
    \includegraphics[width=\textwidth]{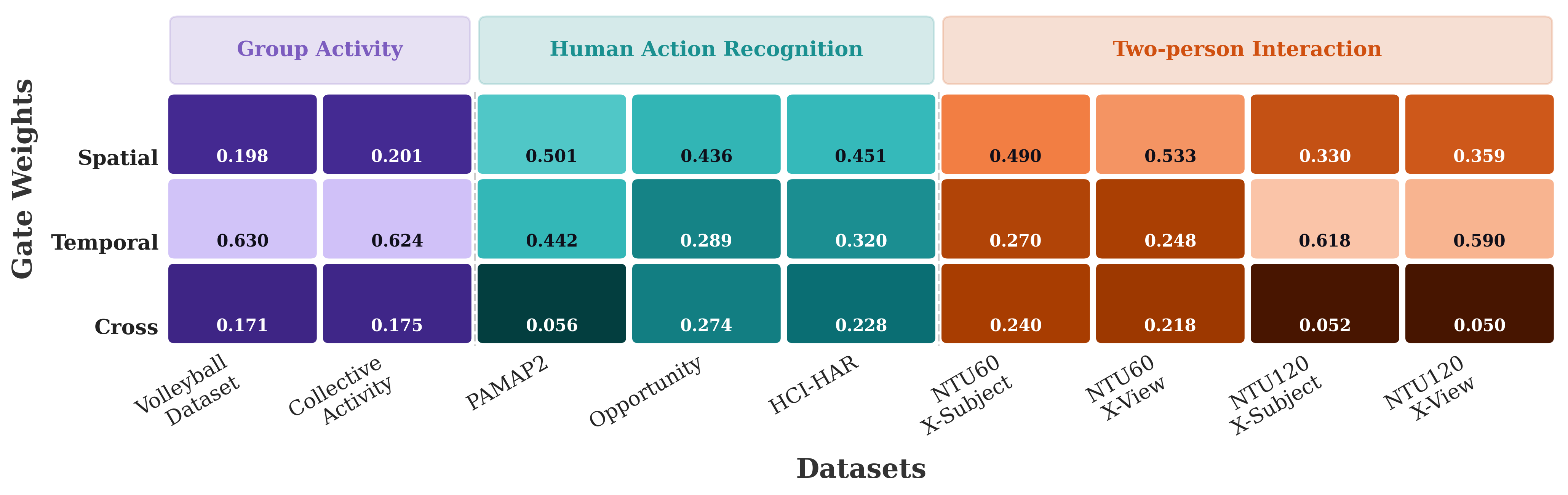}
    \caption{
    Average gated fusion weights across datasets and domains. The learned fusion weights reveal distinct interaction preferences across tasks. Video-based group activity datasets assign larger weights to temporal interactions, sensor-based HAR datasets rely on a more balanced combination of spatial and temporal reasoning, while skeleton-based two-person interaction datasets strongly emphasize temporal dependencies with comparatively smaller cross-interaction contributions. These results indicate that the proposed structured block adaptively reweights spatial, temporal, and cross spatio-temporal interactions according to the characteristics of each domain.
    }
    \label{fig:gate_weights}
\end{figure*}

\paragraph{Analysis of gated fusion weights.}
Fig.~\ref{fig:gate_weights} visualizes the average learned fusion weights of the spatial, temporal, and cross-interaction branches across all datasets. Despite using the same architecture across domains, the model automatically adapts the relative importance of each interaction type. Group activity datasets place greater emphasis on temporal reasoning, likely due to the importance of coordinated motion patterns over time. Sensor-based HAR datasets exhibit a more balanced weighting between spatial and temporal branches, reflecting the coupled nature of sensor-channel and temporal dynamics. In contrast, two-person interaction datasets strongly prioritize temporal interactions, suggesting that motion evolution is the dominant cue for interaction recognition. Across all datasets, the cross-interaction branch consistently receives non-zero weights, demonstrating that explicit spatial--temporal coupling contributes complementary relational information even when it is not the dominant component.

\subsection{Extended Results}

To ensure robust evaluation, we repeat our method over multiple independent runs using different random seeds and report the results as mean$\pm$standard deviation. This captures variations arising from random initialization, data shuffling, and stochastic optimization. For prior methods, we report the values from the corresponding papers.

\captionof{table}[h]{
Comparison of accuracy and F1 score on PAMAP2, Opportunity, and UCI-HAR datasets.
Values are reported as mean$\pm$standard deviation. Red denotes the best result, blue denotes the second-best result, and bold denotes the third-best result in each column.
}
\label{tab:har_comparison}

\setlength{\tabcolsep}{5pt}
\renewcommand{\arraystretch}{1.15}
\resizebox{\linewidth}{!}{%
\begin{tabular}{@{}lcccccc@{}}
\toprule
\textbf{Method}
& \multicolumn{2}{c}{\textbf{PAMAP2}}
& \multicolumn{2}{c}{\textbf{Opportunity}}
& \multicolumn{2}{c}{\textbf{UCI-HAR}} \\
\cmidrule(lr){2-3} \cmidrule(lr){4-5} \cmidrule(lr){6-7}
& \textbf{Accuracy (\%)}
& \textbf{F1 Score (\%)}
& \textbf{Accuracy (\%)}
& \textbf{F1 Score (\%)}
& \textbf{Accuracy (\%)}
& \textbf{F1 Score (\%)} \\
\midrule
Layer-wise CNN~\cite{teng2020layerwisecnn} 
& 92.97 & 93.03 & 81.00 & 80.55 & 96.98 & 96.97 \\

LSTM-CNN~\cite{xia2020lstmcnn}       
& -- & -- & 92.63 & 92.63 & 95.78 & 95.18 \\

Attn-HAR~\cite{mahmud2020attnhar}       
& 96.00 & 96.00 & 67.00 & 42.00 & -- & -- \\

DanHAR~\cite{gao2021danhar}         
& 93.16 & -- & 82.75 & -- & -- & -- \\

Contrast-HAR~\cite{cheng2023contrasthar}   
& 93.22 & 92.97 & -- & -- & 98.00 & 98.00 \\

AFVF~\cite{nguyen2024afvf}           
& \textbf{96.72} & \textbf{96.65} & -- & -- & \textbf{98.61} & \textbf{98.65} \\

MLCNNwav~\cite{dahou2024mlcnnwav}       
& -- & -- & \textbf{93.19} & \textbf{93.30} & 95.52 & 96.11 \\

STDual-X~\cite{chandirakumar2025stdualx}      
& \textcolor{blue}{97.02} 
& \textcolor{blue}{97.04} 
& \textcolor{blue}{95.43} 
& \textcolor{blue}{94.99} 
& \textcolor{blue}{98.80} 
& \textcolor{blue}{98.87} \\

\midrule
\textbf{Ours}  
& \textcolor{red}{99.76$\pm$0.04} 
& \textcolor{red}{99.68$\pm$0.05} 
& \textcolor{red}{99.71$\pm$0.06} 
& \textcolor{red}{99.58$\pm$0.07} 
& \textcolor{red}{99.02$\pm$0.08} 
& \textcolor{red}{99.14$\pm$0.06} \\
\bottomrule
\end{tabular}%
}

\captionof{table}[h]{
Comparison with previous methods on the interaction subsets of NTU RGB+D 60 and NTU RGB+D 120. 
Values for Ours are reported as mean$\pm$standard deviation. 
Red denotes the best result, blue denotes the second-best result, and bold denotes the third-best result in each column. 
For accuracy columns, higher is better; for Params and FLOPs, lower is better.
}
\label{tab:ntu_interaction_comparison}

\setlength{\tabcolsep}{4pt}
\renewcommand{\arraystretch}{1.08}
\resizebox{\linewidth}{!}{%
\begin{tabular}{@{}lcccccc@{}}
\toprule
\textbf{Method}
& \multicolumn{2}{c}{\textbf{NTU RGB+D 60}}
& \multicolumn{2}{c}{\textbf{NTU RGB+D 120}}
& \textbf{Param.(M)}
& \textbf{FLOPs(G)} \\
\cmidrule(lr){2-3} \cmidrule(lr){4-5}
& \textbf{Cross-subject (\%)}
& \textbf{Cross-view (\%)}
& \textbf{Cross-subject (\%)}
& \textbf{Cross-set (\%)}
&  &  \\
\midrule
SGN~\cite{zhang2020sgn}          
& 89.0 & 94.5 & 79.2 & 81.5 & \textcolor{blue}{0.62} & -- \\

3s RA-GCN~\cite{song2021ragcn}    
& 87.3 & 93.6 & 81.1 & 82.7 & 6.21 & -- \\

LSTM-IRN~\cite{perez2022irn}     
& 90.5 & 93.5 & 77.7 & 79.6 & 5.76 & -- \\

Lee et al.~\cite{lee2022interacting}   
& 89.6 & 90.7 & -- & -- & -- & -- \\

CoAGCN~\cite{hedegaard2023continual}       
& 84.1 & 92.6 & 84.0 & 85.3 & 3.47 & \textcolor{blue}{0.17} \\

CoS-TR~\cite{hedegaard2023continual}       
& 86.3 & 92.4 & 84.8 & 86.1 & 3.09 & \textcolor{red}{0.15} \\

3S-AimCLR++~\cite{guo2024aimclr}  
& 87.1 & 93.0 & 82.5 & 83.2 & -- & -- \\

AS-GCN~\cite{li2019asgcn}       
& 89.3 & 93.0 & 82.9 & 83.7 & 6.99 & -- \\

ST-GCN~\cite{yan2018stgcn}       
& 83.3 & 88.7 & 80.2 & 79.0 & 3.14 & 16.73 \\

2s-AGCN~\cite{shi2019agcn}      
& -- & -- & 86.1 & 88.1 & 6.94 & -- \\

ISTA-Net~\cite{wen2023istanet}     
& -- & -- & \textcolor{red}{90.5} & \textcolor{red}{91.7} & 6.22 & 68.18 \\

IGFormer~\cite{pang2022igformer}     
& \textbf{93.6} & \textcolor{blue}{96.5} & 86.5 & 85.4 & -- & -- \\

SkeleTR~\cite{duan2023skeletr}      
& \textcolor{red}{94.9} & \textcolor{red}{97.7} & 88.3 & 87.8 & 3.82 & 7.30 \\

2D-DWT~\cite{wu2025skeleton2ddwt}       
& \textbf{93.6} & 94.6 & \textbf{88.7} & \textcolor{blue}{89.3} & \textcolor{red}{0.53} & 1.89 \\

\midrule
\textbf{Ours} 
& \textcolor{blue}{93.62$\pm$0.5} 
& \textbf{95.46$\pm$0.4} 
& \textcolor{blue}{88.94$\pm$0.7} 
& \textbf{88.26$\pm$0.6} 
& \textbf{1.76} 
& \textbf{0.49} \\
\bottomrule
\end{tabular}%
}

\begin{table}[h]
\centering
\caption{
Comparison with previous methods on the Volleyball and Collective Activity datasets.
Values for Ours are reported as mean$\pm$standard deviation.
Red denotes the best result, blue denotes the second-best result, and bold denotes the third-best result in each metric column.
For accuracy metrics, higher is better; for Params and GFLOPs, lower is better.
}
\label{tab:volleyball_collective_acc}
\footnotesize
\setlength{\tabcolsep}{4pt}
\renewcommand{\arraystretch}{1.08}
\resizebox{\linewidth}{!}{%
\begin{tabular}{@{}lccccccc@{}}
\toprule
\textbf{Method}
& \multicolumn{5}{c}{\textbf{Volleyball Dataset}}
& \multicolumn{2}{c}{\textbf{Collective Activity Dataset}} \\
\cmidrule(lr){2-6} \cmidrule(lr){7-8}
& \textbf{Backbone}
& \textbf{MCA}
& \textbf{Merged MCA}
& \textbf{Params}
& \textbf{GFLOPs}
& \textbf{Backbone}
& \textbf{MPCA} \\
\midrule
PCTDM~\cite{yan2018pctdm}         
& ResNet-18    
& 90.3 
& 94.3 
& --    
& --    
& AlexNet      
& 92.2  \\

StagNet~\cite{qi2018stagnet}       
& VGG-16       
& 89.3 
& --   
& --    
& --    
& VGG-16       
& 89.1  \\

ARG~\cite{wu2019arg}           
& ResNet-18    
& 91.1 
& \textbf{95.1} 
& \textbf{49.5M} 
& \textbf{307G}  
& ResNet-18    
& 92.3  \\

CRM~\cite{azar2019crm}           
& I3D          
& 92.1 
& --   
& --    
& --    
& I3D          
& 94.2  \\

HiGCIN~\cite{yan2020higcin}        
& ResNet-18    
& 91.4 
& --   
& --    
& --    
& ResNet-18    
& 93.0  \\

AT~\cite{gavrilyuk2020actortransformer}            
& ResNet-18    
& 90.0 
& 94.0 
& --    
& --    
& --           
& --    \\

SACRF~\cite{pramono2020sacrf}         
& ResNet-18    
& 90.7 
& 92.7 
& 53.7M 
& 339G  
& --           
& --    \\

DIN~\cite{yuan2021din}           
& ResNet-18    
& 93.1 
& \textcolor{red}{95.6} 
& \textcolor{blue}{26.0M} 
& \textcolor{blue}{304G}  
& ResNet-18    
& \textbf{95.3}  \\

TCE+STBiP~\cite{yuan2021tce}     
& VGG-16       
& \textbf{94.1} 
& --   
& --    
& --    
& Inception-v3 
& 95.1  \\

GroupFormer~\cite{li2021groupformer}   
& Inception-v3 
& \textbf{94.1} 
& --   
& --    
& --    
& --           
& --    \\

Dual-AI (RGB)~\cite{han2022dualai} 
& Inception-v3 
& \textcolor{blue}{94.4} 
& --   
& --    
& --    
& Inception-v3 
& \textcolor{blue}{96.5}  \\

\midrule
\textbf{Ours} 
& \textbf{ResNet-18} 
& \textcolor{red}{94.6$\pm$0.2} 
& \textcolor{blue}{95.3$\pm$0.2} 
& \textcolor{red}{13.38M} 
& \textcolor{red}{18G}
& \textbf{ResNet-18} 
& \textcolor{red}{97.47$\pm$0.3} \\
\bottomrule
\end{tabular}%
}
\end{table}
\newpage

\end{document}